\documentclass[11pt]{article}

\usepackage{acl}

\usepackage{times}
\usepackage{latexsym}
\usepackage[T1]{fontenc}
\usepackage[utf8]{inputenc}
\usepackage{microtype}
\usepackage{booktabs}
\usepackage{xurl}
\usepackage{tabularx}
\usepackage{etoolbox}
\apptocmd{\thebibliography}{\sloppy}{}{}
\usepackage{amsmath}
\usepackage{amssymb}
\usepackage{multirow}
\usepackage{inconsolata}
\usepackage{tikz}
\usetikzlibrary{positioning,arrows.meta,fit,backgrounds,calc}
\usepackage{graphicx}
\usepackage{comment}

\title{HelaBERT: Enhancing Sinhala Language Understanding with Dual Pooling Classification Head}

\author{Thisen Ekanayake \quad Nisansa de Silva \\
  Department of Computer Science \& Engineering \\
  University of Moratuwa \\
  \texttt{\{thisene.23,NisansaDdS\}@cse.mrt.ac.lk}}

\begin{document}
\raggedbottom
\setlength{\emergencystretch}{3em}
\maketitle

\begin{abstract}
We present HelaBERT, a family of two BERT-based masked language models
pre-trained from scratch on approximately 1 billion tokens of Sinhala
text sourced from MADLAD-400, CulturaX, and a custom corpus comprising
news articles, Sinhala Wikipedia, and web crawl data.
HelaBERT-Small ($\sim$23.3M parameters, 6 layers) and HelaBERT-Large
($\sim$110M parameters, 12 layers) both use a SentencePiece Unigram
tokenizer (vocabulary size 32,000) tailored to Sinhala's agglutinative
morphology and complex script.
We evaluate both models on four downstream Sinhala text
classification tasks: news category classification, news source
classification, sentiment analysis, and writing style classification,
using 5 independent seed runs with stratified 80/20 train/test splits.
We additionally propose a dual pooling classification head and evaluate it
systematically across all four tasks, finding consistent improvements on
sentiment analysis and a moderate gain on news category classification
for HelaBERT-Small, while the standard \texttt{[CLS]}-linear
head remains competitive on news source classification, a headline-level
task with short average input length.
We release both models to support further research in Sinhala NLP.
\end{abstract}

\section{Introduction}
\label{sec:intro}

Sinhala is an Indo-Aryan language spoken by approximately 17 million
people in Sri Lanka, characterized by a complex abugida script and rich
agglutinative morphology. Despite its regional importance, Sinhala
remains severely under-resourced in NLP: pre-trained language models,
annotated corpora, and benchmarks are scarce. While multilingual models
such as mBERT \cite{devlin-etal-2019-bert} and XLM-R
\cite{conneau-etal-2020-unsupervised} provide some coverage, they
allocate limited capacity to low-resource languages, and dedicated
monolingual models consistently outperform them on downstream tasks
\cite{martin-etal-2020-camembert,delobelle-etal-2020-robbert}.

We introduce \textbf{HelaBERT}, a family of two BERT-based masked
language models pre-trained from scratch on $\sim$1 billion tokens of
Sinhala text, using a SentencePiece Unigram tokenizer designed for
Sinhala's agglutinative morphology and complex script. We fine-tune
both models on four Sinhala text classification tasks --- news
category, news source, sentiment, and writing style --- and compare
against existing multilingual and monolingual baselines under a
common evaluation protocol. Beyond the standard \texttt{[CLS]}-linear
classification head, we propose a \textbf{dual pooling} head that
lets the \texttt{[CLS]} token and the full token sequence attend to
each other, and we analyze systematically when this richer
interaction helps and when it does not. Our main contributions are:
\begin{enumerate}
  \item Two BERT-based models: \raisebox{-2.2pt}{\includegraphics[scale=0.09]{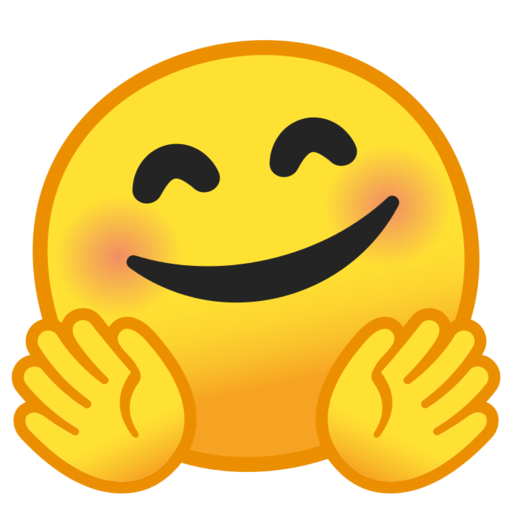}} \href{https://huggingface.co/ThisenEkanayake/HelaBERT}{\texttt{HelaBERT-Small}} and \raisebox{-2.2pt}{\includegraphics[scale=0.09]{images/huggingface.png}} \href{https://huggingface.co/ThisenEkanayake/HelaBERT\_Large}{\texttt{HelaBERT\_Large}}, with a SentencePiece Unigram tokenizer tailored to Sinhala
        morphology;
  \item A fine-tuning evaluation on four Sinhala classification
        benchmarks using 5 independent seed runs following the
        methodology of \citet{dhananjaya-etal-2022-bertifying}, enabling
        direct comparison with SinBERT (\citet{dhananjaya-etal-2022-bertifying}) and other baselines; and
  \item A dual pooling classification head evaluated across all four
        tasks, yielding consistent gains on sentiment analysis
        ($+$3.9--5.6 macro-F$_1$ points) and a moderate improvement on
        news category classification for HelaBERT-Small ($+$3.1 points),
        while demonstrating that the standard head remains competitive
        on very short-input tasks such as news source classification.
\end{enumerate}

Section~\ref{sec:related} situates HelaBERT relative to prior
multilingual and Sinhala-specific models; Sections~\ref{sec:pretraining}--\ref{sec:finetuning}
describe pre-training and fine-tuning; Section~\ref{sec:results}
compares HelaBERT against baselines; Section~\ref{sec:coattn}
introduces and analyzes the co-attention head; and
Section~\ref{sec:discussion} discusses broader implications and
limitations.

\section{Related Work}
\label{sec:related}

\subsection{Multilingual and Monolingual Pre-trained Language Models}

BERT \cite{devlin-etal-2019-bert} established masked language modelling
as the dominant NLP pre-training paradigm. Multilingual extensions such
as mBERT and XLM-R \cite{conneau-etal-2020-unsupervised} provide broad
language coverage but allocate limited capacity to low-resource
languages; dedicated monolingual models consistently outperform them
\cite{dhananjaya-etal-2022-bertifying,martin-etal-2020-camembert}.
Language-specific models for French (CamemBERT;
\citealt{martin-etal-2020-camembert}), Arabic (AraBERT;
\citealt{antoun-etal-2020-arabert}), Vietnamese (PhoBERT;
\citealt{nguyen-tuan-nguyen-2020-phobert}), and Indic languages
(IndicBERT; \citealt{kakwani-etal-2020-indicnlpsuite}) confirm this
pattern across diverse, morphologically rich settings. HelaBERT follows
this line of work for Sinhala.

\subsection{Pre-trained Language Models for Sinhala}

\paragraph{SinBERT.}
\citet{dhananjaya-etal-2022-bertifying} pre-trained two RoBERTa-based
\cite{liu2019roberta} monolingual models --- SinBERT-Small and
SinBERT-Large --- and evaluated them on four classification benchmarks,
introducing the evaluation datasets and protocol reused in this work.
HelaBERT competes directly on the same tasks using the same evaluation
methodology, while contrasting in backbone (BERT vs.\ RoBERTa) and
tokenizer (SentencePiece Unigram vs.\ BPE).

\paragraph{SinLlama.}
\citet{aravinda2025sinllama} introduced the first decoder-based
Sinhala LLM via continual pre-training of Llama-3-8B
\cite{grattafiori2024llama}. HelaBERT and SinLlama are
complementary: a lightweight encoder for classification and sequence
labelling versus a generative model for instruction-following.

\subsection{Sinhala News Category Classification}

Prior work on topical categorization of Sinhala news includes an
early Na\"ive Bayes and SVM system and an LDA-based approach that
builds Sinhala news topic hierarchies for categorization
\citep{desilva2026surveypubliclyavailablesinhala}. The benchmark we
use, however, is the news category dataset introduced by
\citet{dhananjaya-etal-2022-bertifying}, which we adopt for direct
comparability with SinBERT.

\subsection{Sinhala News Source Classification}

News source identification for Sinhala has mostly been studied
alongside the related task of misinformation detection, including an
ontology-based approach to fake news detection and a credibility-tagged
Sinhala news dataset \citep{desilva2026surveypubliclyavailablesinhala}.
We evaluate on the news source dataset released by
\citet{dhananjaya-etal-2022-bertifying}, a headline-level task with
short average input length.

\subsection{Sinhala Writing Style Classification}

Writing style classification for Sinhala has been explored
directly, including a character-level model for identifying student
authors and a dedicated writing-style identification dataset covering
Romanized Sinhala text \citep{desilva2026surveypubliclyavailablesinhala}.
We use the writing style classification dataset of
\citet{dhananjaya-etal-2022-bertifying}, on which prior BERT-based
models already report near-ceiling performance.

\subsection{Sinhala Sentiment Analysis}

Pre-transformer work established strong baselines using Word2Vec and
fastText embeddings \cite{senevirathne2020sentiment} and
hierarchical attention and capsule networks
\cite{10.1145/3445035}.
HelaBERT extends this line
of work with a dual pooling classification head that operates over the intra-sequence
interaction between the \texttt{[CLS]} token and the remaining token
representations.

\section{HelaBERT Pre-training}
\label{sec:pretraining}

We pre-train two models --- HelaBERT-Small and HelaBERT-Large --- sharing
the same corpus, tokenizer, and MLM objective, but differing in
architecture scale and training configuration.

\subsection{Pre-training Data}
\label{subsec:pretraining-data}

HelaBERT-Small was pre-trained on approximately \textbf{900 million tokens} and HelaBERT-Large on approximately \textbf{1.1 billion tokens} of Sinhala text sourced from three corpora:

\begin{itemize}
  \item \textbf{MADLAD-400} \cite{kudugunta2023madlad}: the Sinhala
        subset of the multilingual document-level dataset.
  \item \textbf{CulturaX} \cite{nguyen-etal-2024-culturax}: the Sinhala
        subset of the cleaned multilingual web corpus.
  \item \textbf{Custom Sinhala Corpus}: a dataset compiled from
        Sinhala Wikipedia, Sinhala news articles, and Sinhala web crawl
        data.
\end{itemize}

\subsection{Data Preprocessing}
\label{subsec:preprocessing}
Raw text was NFC-normalized, invisible Unicode characters removed (retaining
ZWJ/ZWNJ for Sinhala ligature rendering), and lines lacking Sinhala script or fewer than five characters discarded. Non-Sinhala
characters were stripped, preserving ASCII digits, punctuation, and ZWJ/ZWNJ.
Repeated punctuation, extra whitespace, unmatched brackets, and date-like
numeric patterns were then normalized before tokenization with the SentencePiece
Unigram model described below.

\subsection{Tokenizer}
\label{subsec:tokenizer}

Both models use a \textbf{SentencePiece Unigram} tokenizer
\cite{kudo-richardson-2018-sentencepiece} with a vocabulary size of 32,000 and a
character coverage of 99.95\%. The tokenizer is trained from scratch on
a $\sim$180M-token subset of the pre-training corpus described in
Section~\ref{subsec:pretraining-data}, rather than adapted from an
existing multilingual vocabulary. Operating directly on raw Unicode
without word-boundary assumptions, it is well-suited to Sinhala's
agglutinative morphology and complex script.

\subsection{Model Architectures}
\label{subsec:model-arch}

Both models follow the standard BERT encoder-only architecture
\cite{devlin-etal-2019-bert} with MLM as the pre-training objective.
Table~\ref{tab:arch} compares the two configurations.

\begin{table}[!htbp]
\centering
\small
\resizebox{\columnwidth}{!}{%
\begin{tabular}{lll}
\toprule
\textbf{Parameter} & \textbf{Small} & \textbf{Large} \\
\midrule
Parameters             & $\sim$23.3M & $\sim$110M  \\
Hidden size            & 384         & 768         \\
Transformer layers     & 6           & 12          \\
Attention heads        & 6           & 12          \\
Intermediate size      & 1,536       & 3,072       \\
Max sequence length    & 512         & 512         \\
Vocabulary size        & 32,000      & 32,000      \\
Activation function    & GELU        & GELU        \\
Position embeddings    & Absolute    & Absolute    \\
Hidden dropout         & 0.1         & 0.1         \\
Attention dropout      & 0.1         & 0.1         \\
\bottomrule
\end{tabular}%
}
\caption{Architecture comparison between HelaBERT-Small and HelaBERT-Large.}
\label{tab:arch}
\end{table}

\paragraph{MLM Objective.}
At each step, 15\% of non-padding tokens are selected: 80\% replaced with
\texttt{[MASK]}, 10\% with a random token, and 10\% left unchanged.
Cross-entropy loss is computed only over masked positions; unmasked
positions are assigned label $-100$ and excluded.

\subsection{Pre-training Configuration}
\label{subsec:pretraining-config}

Both models were pre-trained using the HuggingFace \texttt{Trainer} API
\cite{wolf-etal-2020-transformers} with a custom \texttt{SimpleMLMCollator}
applying masking per batch at runtime.\footnote{\small Pre-training code:
\url{https://anonymous.4open.science/r/HelaBERT-Train}. Code was developed
with assistance from Claude Code (Anthropic).} Full hyperparameters are listed in
Table~\ref{tab:pretrain-config} (Appendix~\ref{sec:appendix-hyperparams}).

\subsection{Pre-training Results}
\label{subsec:pretraining-results}

\begin{figure}[t]
  \includegraphics[width=\columnwidth]{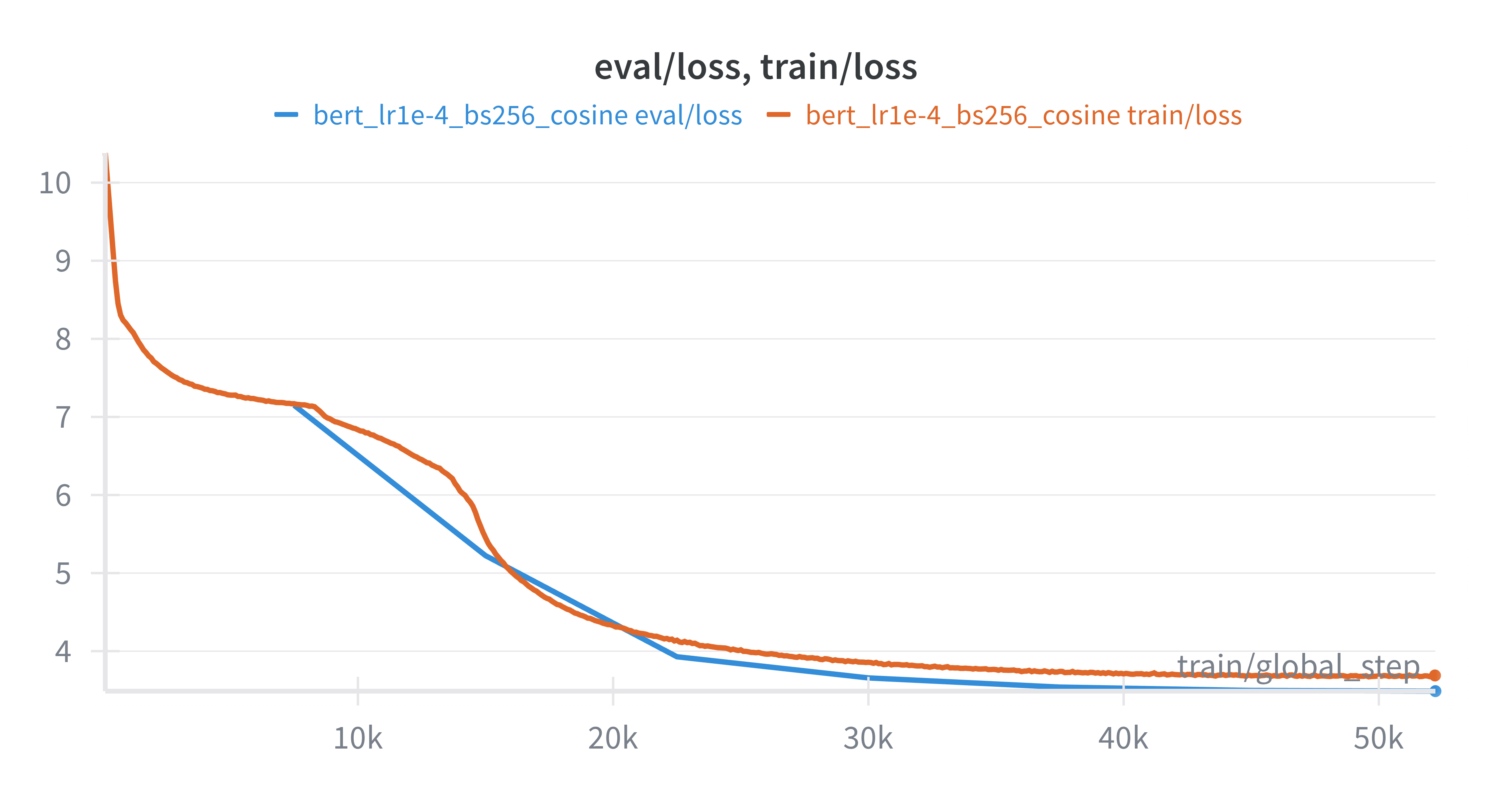}
  \caption{Training and validation loss curves for HelaBERT-Small.}
  \label{fig:loss-curve-small}
\end{figure}

HelaBERT-Small trained for 2 epochs ($\sim$52,000 steps). Training loss
decreased from $\sim$10.0 to 3.69 and validation loss from $\sim$7.0 to
3.49, with no significant overfitting observed
(Figure~\ref{fig:loss-curve-small}).

\begin{figure}[t]
  \includegraphics[width=\columnwidth]{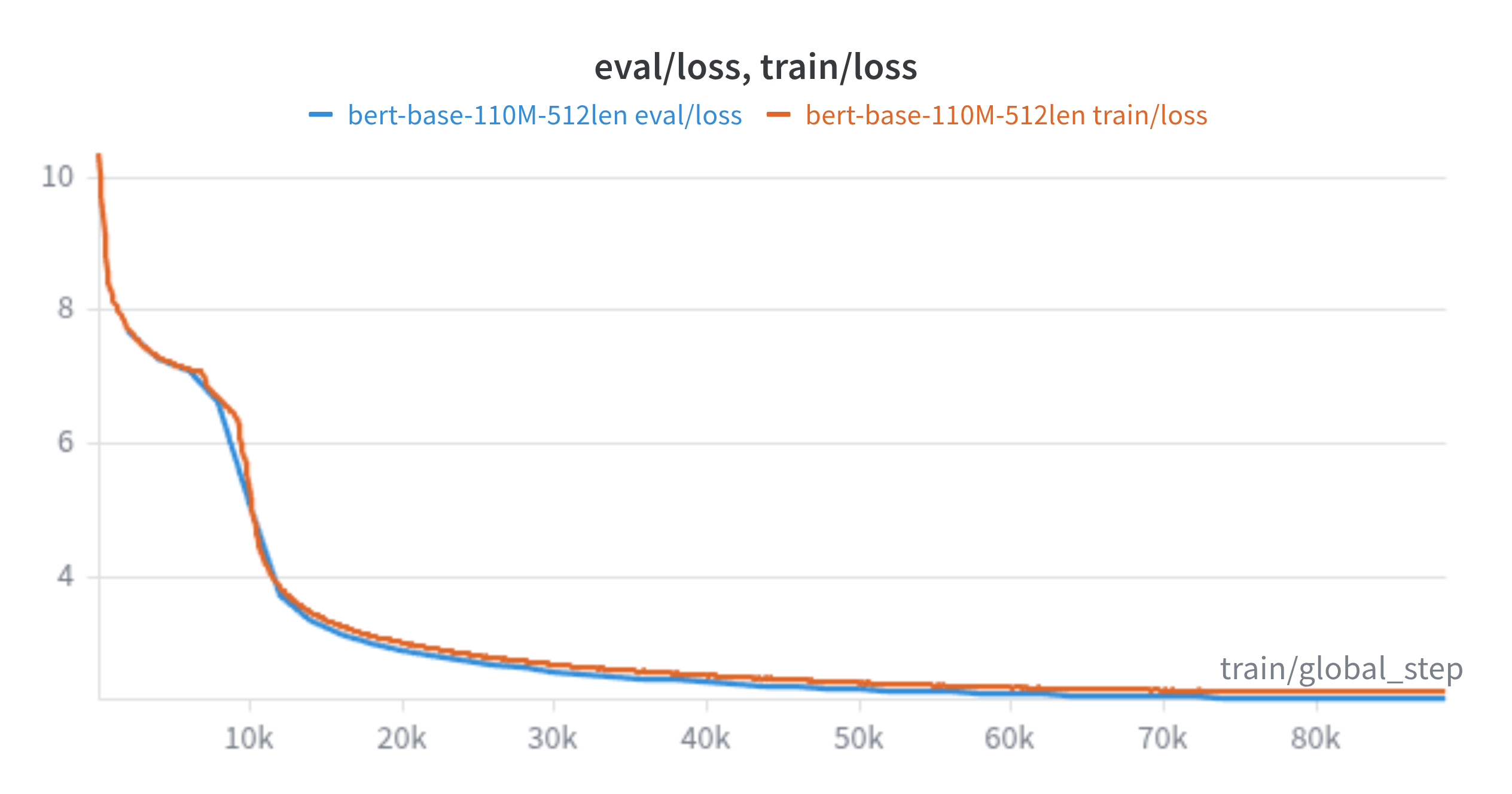}
  \caption{Training and validation loss curves for HelaBERT-Large.}
  \label{fig:loss-curve-large}
\end{figure}

HelaBERT-Large trained for 6 epochs ($\sim$90,700 steps). Training loss
decreased from $\sim$10.3 to 2.26 and validation loss from $\sim$7.5 to
2.17, again with no significant overfitting
(Figure~\ref{fig:loss-curve-large}). The substantially lower
final loss relative to HelaBERT-Small reflects the larger model's
representational capacity. Final metrics for both models are reported in
Table~\ref{tab:pretrain-results}.

\begin{table}[!htbp]
\centering
\small
\resizebox{\columnwidth}{!}{%
\begin{tabular}{lll}
\toprule
\textbf{Metric}        & \textbf{Small} & \textbf{Large} \\
\midrule
Final train loss       & 3.69  & 2.26  \\
Final eval loss        & 3.49  & 2.17  \\
Total training steps   & $\sim$52,000 & $\sim$90,700 \\
\bottomrule
\end{tabular}%
}
\caption{Pre-training results for HelaBERT-Small and HelaBERT-Large.}
\label{tab:pretrain-results}
\end{table}

\subsection{Hardware and Environmental Impact}
\label{subsec:hardware}
HelaBERT-Small was trained on a single NVIDIA RTX 4060 (8\,GB, 55\,W) for
$\sim$16 hours, yielding an estimated footprint of $\approx$0.29\,kg
CO$_2$eq (Sri Lanka grid: 0.329\,kg CO$_2$/kWh \cite{owid-carbon-intensity}).
HelaBERT-Large was trained on an AMD Instinct MI300X (192\,GB, 700\,W) via
DigitalOcean (Atlanta) for $\sim$22.5 hours, giving $\approx$6.09\,kg
CO$_2$eq (US-SRSO grid: 0.384\,kg CO$_2$/kWh \cite{owid-carbon-intensity}).

\section{Fine-tuning Datasets}
\label{sec:finetuning-datasets}

We fine-tune and evaluate both HelaBERT models on four Sinhala text
classification tasks, following the methodologies used by
\citet{dhananjaya-etal-2022-bertifying} to enable direct comparison.

\subsection{News Category Classification}

Sinhala news sentences across five categories: \textit{political},
\textit{business}, \textit{technology}, \textit{sports}, and
\textit{entertainment}. Originally introduced by
\citet{de2015sinhala} and cleaned by
\citet{dhananjaya-etal-2022-bertifying}.\footnote{\small\url{https://huggingface.co/datasets/NLPC-UOM/Sinhala-News-Category-classification}}

\subsection{News Source Classification}

Sinhala news headlines from nine online sources. Derived from
\citet{sachintha2021exploiting} and processed by
\citet{dhananjaya-etal-2022-bertifying}.\footnote{\small\url{https://huggingface.co/datasets/NLPC-UOM/Sinhala-News-Source-classification}}

\subsection{Sentiment Analysis}
\label{subsec:sentiment-dataset}

We use a publicly available Sinhala sentiment dataset with three labels:
\textsc{positive}, \textsc{negative}, and \textsc{neutral}.\footnote{\small\url{https://huggingface.co/datasets/sinhala-nlp/sinhala-sentiment-analysis}}

We note that \citet{dhananjaya-etal-2022-bertifying} and all prior
baselines were evaluated on a different four-class sentiment dataset that
also includes a \textsc{conflict} label; that dataset is no longer
publicly available. Our sentiment results therefore use a different,
publicly accessible three-class dataset and are \textbf{not directly
comparable} to those baselines on this task specifically. We report them
alongside the other baselines for reference, with this distinction
clearly noted.

\subsection{Writing Style Classification}

Longer-form Sinhala text across four styles: \textsc{news},
\textsc{academic}, \textsc{creative}, and \textsc{blog}. Originally compiled by \citet{upeksha2015corpus} and released by
\citet{dhananjaya-etal-2022-bertifying}.\footnote{\small\url{https://huggingface.co/datasets/NLPC-UOM/Writing-style-classification}}

\subsection{Dataset Overview}

Table~\ref{tab:datasets} summarises statistics for all four datasets.

\begin{table}[!htbp]
\centering
\small
\resizebox{\columnwidth}{!}{%
\begin{tabular}{lrrrr}
\toprule
\textbf{Dataset} & \textbf{Classes} & \textbf{Train} & \textbf{Test} & \textbf{Avg.\ W} \\
\midrule
News Category  & 5 & 2,596  & 640   & 23.8  \\
News Source    & 9 & 18,280 & 4,571 & 8.3   \\
Sentiment  & 3 & 2,049  & 513   & 16.8  \\
Writing Style  & 4 & 10,008 & 2,462 & 182.6 \\
\bottomrule
\end{tabular}%
}
\caption{Summary statistics of the four fine-tuning datasets. Avg.\ Words
computed over the training split.}
\label{tab:datasets}
\end{table}

\section{Fine-tuning Methodology}
\label{sec:finetuning}

We fine-tune both HelaBERT models on the four tasks described above.\footnote{\small Fine-tuning code:
\url{https://anonymous.4open.science/r/HelaBERT_Analysis}. Code was developed
with assistance from Claude Code (Anthropic).}
All experiments follow the evaluation setup of
\citet{dhananjaya-etal-2022-bertifying}: 5 independent training runs
with different random seeds, a stratified 80/20 train/test split, and
macro-F$_1$ as the primary evaluation metric. We report results from the
best-performing run (highest test macro-F$_1$) for each task and model.

\subsection{Standard Classification Head}
\label{sec:cls-head}

For all four tasks, the fine-tuning architecture consists of the
pre-trained HelaBERT backbone followed by a classification head applied
to the \texttt{[CLS]} token representation
$\mathbf{h}_{\texttt{CLS}} \in \mathbb{R}^H$:
\begin{equation}
  \hat{y} = \mathrm{Linear}(\mathrm{Dropout}(\mathbf{h}_{\texttt{CLS}})).
\end{equation}
The linear layer maps from hidden size $H$ to the number of classes, with
dropout probability 0.1.

\subsection{Training Configuration}
\label{sec:training}

All models are trained using the HuggingFace \texttt{Trainer} API
\cite{wolf-etal-2020-transformers} with AdamW optimization, linear
learning rate scheduling with a 6\% warmup ratio, weight decay 0.01,
batch size 16, and FP16 mixed precision. The best run is selected by
highest test macro-F$_1$ across five seeds
(42, 123, 456, 789, 1024). Per-task hyperparameters are listed in
Table~\ref{tab:hyperparams}.

\begin{table}[!htbp]
\centering
\small
\resizebox{\columnwidth}{!}{%
\begin{tabular}{lcccc}
\toprule
& \textbf{News Cat.} & \textbf{News Src.} & \textbf{Senti.} & \textbf{Style} \\
\midrule
LR (Small)    & 1e-5 & 5e-5 & 3e-5 & 1e-5 \\
LR (Large)    & 3e-5 & 5e-5 & 5e-6 & 1e-5 \\
Epochs (Small)& 10   & 3    & 6    & 3    \\
Epochs (Large)& 3    & 3    & 10   & 3    \\
Max seq. len. & 512  & 512  & 512  & 512  \\
Batch size    & \multicolumn{4}{c}{16} \\
Weight decay  & \multicolumn{4}{c}{0.01} \\
Warmup ratio  & \multicolumn{4}{c}{0.06} \\
LR schedule   & \multicolumn{4}{c}{Linear} \\
\bottomrule
\end{tabular}%
}
\caption{Fine-tuning hyperparameters per task. LR = learning rate.}
\label{tab:hyperparams}
\end{table}

\section{Results: Comparison with Baselines}
\label{sec:results}

Table~\ref{tab:comparison} reports macro-F$_1$ scores for HelaBERT-Small
and HelaBERT-Large alongside baseline results reported by
\citet{dhananjaya-etal-2022-bertifying}. The baselines include LaBSE,
LASER, XLM-R (base and large), SinBERT$_O$, SinhalanBERT$_O$,
SinBERT-Small, and SinBERT-Large, all evaluated on the same News
Category, News Source, and Writing Style datasets. As noted in
Section~\ref{subsec:sentiment-dataset}, sentiment results for HelaBERT
are \textbf{not directly comparable} to the baselines due to the use of
a different three-class dataset; they are included for completeness with
an explicit marker.

\begin{table*}[!htbp]
\centering
\small
\begin{tabular}{lcccc}
\toprule
\textbf{Model} &
\textbf{Sentiment$^\dagger$} &
\textbf{News Source} &
\textbf{News Category} &
\textbf{Writing Style} \\
\midrule
Baseline (majority class)      & 59.42 (w-F$_1$) & ---   & ---   & ---   \\
LaBSE                          & 20.63           & 11.85 & 24.09 & ---   \\
LASER                          & 54.07           & 28.84 & 48.54 & 87.06 \\
XLM-R$_{\text{base}}$          & 58.08           & 58.29 & 85.12 & 96.89 \\
XLM-R$_{\text{large}}$         & 60.45           & 61.84 & 89.54 & 98.41 \\
SinBERT$_O$                    & 50.83           & 57.22 & 78.07 & 93.84 \\
SinhalanBERT$_O$               & 49.71           & 57.34 & 82.73 & 94.10 \\
SinBERT-Small                  & 53.85           & 60.42 & 84.75 & 95.00 \\
SinBERT-Large                  & 54.08           & 60.51 & 85.19 & 95.49 \\
\midrule
\textbf{HelaBERT-Small} & \textbf{65.34}  & \textbf{60.16} & \textbf{85.97} & \textbf{95.92} \\
\textbf{HelaBERT-Large} & \textbf{64.60}  & \textbf{63.65} & \textbf{90.38} & \textbf{97.73} \\
\bottomrule
\end{tabular}
\caption{Macro-F$_1$ (\%) comparison on four Sinhala text classification
tasks. All baseline results are taken from \citet{dhananjaya-etal-2022-bertifying}.
$^\dagger$Sentiment results for HelaBERT models are evaluated on a
different publicly available 3-class dataset (positive/negative/neutral);
all other models used a 4-class dataset (including \textsc{conflict})
that is no longer publicly available. These columns are \textbf{not
directly comparable}.}
\label{tab:comparison}
\end{table*}

\paragraph{News Category.}
HelaBERT-Large achieves 90.38\% macro-F$_1$, surpassing XLM-R-large
(89.54\%) by 0.8 points and outperforming all SinBERT variants by a
margin of 5.2--5.6 points over SinBERT-Large (85.19\%) and
SinBERT-Small (84.75\%) respectively.
HelaBERT-Small (85.97\%) outperforms both SinBERT-Small (84.75\%) and
SinBERT-Large (85.19\%) by 1.2 and 0.8 points respectively.

\paragraph{News Source.}
HelaBERT-Large (63.65\%) outperforms all SinBERT models and XLM-R models,
and exceeds SinBERT-Large by 3.1 points. HelaBERT-Small (60.16\%)
is comparable to SinBERT-Small (60.42\%), falling marginally short
by 0.3 points. News source is the most
challenging task, reflecting stylistic overlap across sources.

\paragraph{Writing Style.}
HelaBERT-Large achieves 97.73\%, falling 0.7 points short of XLM-R-large
(98.41\%) but outperforming all SinBERT models by a clear margin of 2.2
points over the best SinBERT-Large (95.49\%).
HelaBERT-Small (95.92\%) similarly outperforms all SinBERT variants
on this task.

\paragraph{Sentiment.}
As discussed, HelaBERT-Small (65.34\%) and HelaBERT-Large (64.60\%) were
evaluated on a three-class dataset. The baseline figures are shown for
reference only and should not be interpreted as performance comparisons.

\section{Dual Pooling Classification Head}
\label{sec:coattn}

Standard fine-tuning for classification uses only the \texttt{[CLS]}
token representation as the sequence summary. We propose a
\textbf{dual pooling} classification head that jointly attends over
\texttt{[CLS]} and the full token sequence within the same input,
producing richer representations that capture both global and local
sequence information. We apply this head to all four tasks on both
HelaBERT-Small and HelaBERT-Large.

\subsection{Dual pooling Architecture}
\label{subsec:coattn-arch}

Given the encoder output, let
$\mathbf{c} = \mathbf{h}_0 \in \mathbb{R}^H$ be the \texttt{[CLS]}
vector and $\mathbf{T} = \{\mathbf{h}_1, \ldots, \mathbf{h}_{T}\}
\in \mathbb{R}^{T \times H}$ be the remaining token representations,
with $\mathbf{m} \in \{0,1\}^T$ the corresponding padding mask.

A shared affinity score is computed for each token position $i$:
\begin{equation}
  a_i = \frac{1}{\sqrt{H}}\, \mathbf{v}^\top
  \tanh\!\bigl(\mathbf{W}_c\,\mathbf{c} + \mathbf{W}_t\,\mathbf{h}_i\bigr),
\end{equation}
where $\mathbf{W}_c, \mathbf{W}_t \in \mathbb{R}^{H \times H}$ and
$\mathbf{v} \in \mathbb{R}^H$ are learned parameters.

We used $\frac{1}{\sqrt{H}}\,$ scaling to keep the affinity scores at a stable magnitude as the hidden dimension increases, preventing the attention scores from becoming excessively large and promoting stable optimization.

\paragraph{Direction 1: \texttt{[CLS]} attends over tokens.}
Softmax attention over real token positions yields an attended
\texttt{[CLS]} representation:
\begin{equation}
  \boldsymbol{\alpha} = \mathrm{softmax}\bigl(\mathbf{a} + (1-\mathbf{m}) \cdot (-10^4)\bigr),
  \quad
  \tilde{\mathbf{c}} = \boldsymbol{\alpha}\,\mathbf{T}.
\end{equation}
Padding positions are masked to $-10^4$ before softmax to prevent
numerical overflow in FP16.

\paragraph{Direction 2: Token sequence attends back to \texttt{[CLS]}.}
A sigmoid gate weighted by the padding mask provides a \texttt{[CLS]}-guided
summary of the token sequence:
\begin{equation}
  \boldsymbol{\beta} = \sigma(\mathbf{a}) \odot \mathbf{m},
  \quad
  \hat{\boldsymbol{\beta}} = \frac{\boldsymbol{\beta}}
      {\sum_j \beta_j + \varepsilon},
  \quad
  \tilde{\mathbf{T}} = \hat{\boldsymbol{\beta}}\,\mathbf{T}.
\end{equation}

\paragraph{Classification head.}
The two attended vectors are layer-normalized, concatenated, and passed
through a two-layer MLP:
\begin{align}
  \mathbf{f} &= \bigl[\mathrm{LN}(\tilde{\mathbf{c}});\,
                \mathrm{LN}(\tilde{\mathbf{T}})\bigr]
                \in \mathbb{R}^{2H}, \\
  \mathbf{h} &= \mathrm{Linear}_{2H \to H}\!\bigl(\mathrm{Dropout}(\mathbf{f})\bigr), \\
  \mathbf{g} &= \mathrm{Dropout}\!\bigl(\mathrm{GELU}(\mathbf{h})\bigr), \\
  \hat{y}    &= \mathrm{Linear}_{H \to C}(\mathbf{g}),
\end{align}
where $C$ is the number of classes.

\begin{figure*}[t]
\centering
\resizebox{0.88\textwidth}{!}{%
\begin{tikzpicture}[
  base/.style={
    rectangle, rounded corners=3pt,
    minimum width=3.2cm, minimum height=0.58cm,
    font=\small, align=center, inner sep=4pt,
    draw=black!70, thick},
  encoder/.style ={base, fill=gray!12},
  split/.style   ={base, fill=teal!14,   minimum width=2.2cm},
  affinity/.style={base, fill=violet!12, minimum width=6.4cm},
  dir1/.style    ={base, fill=blue!10,   minimum width=2.8cm},
  dir2/.style    ={base, fill=orange!12, minimum width=2.8cm},
  combine/.style ={base, fill=violet!12, minimum width=6.4cm},
  mlp/.style     ={base, fill=purple!10, minimum width=7.2cm},
  output/.style  ={base, fill=yellow!15, minimum width=2.8cm,
                   rounded corners=8pt},
  arr/.style={-{Stealth[length=4pt,width=3pt]}, thick, black!65},
  darr/.style={-{Stealth[length=4pt,width=3pt]}, thick,
               black!40, densely dashed},
  lbl/.style={font=\scriptsize, fill=white, inner sep=1pt},
]
\node[encoder] (enc) {HelaBERT Encoder};

\node[split, below left=1.5cm and 0.8cm of enc]  (cls)
    {\texttt{[CLS]} vector\\[1pt]\scriptsize$\mathbf{c}\in\mathbb{R}^{H}$};
\node[split, below=1.5cm of enc] (tok)
    {Token sequence\\[1pt]\scriptsize$\mathbf{T}\in\mathbb{R}^{T\times H}$};
\node[split, below right=1.5cm and 0.8cm of enc] (msk)
    {Padding mask\\[1pt]\scriptsize$\mathbf{m}\in\{0,1\}^{T}$};

\draw[arr] (enc.south) -- ++(0,-0.2) -| (cls.north);
\draw[arr] (enc.south) -- (tok.north);
\draw[arr] (enc.south) -- ++(0,-0.2) -| (msk.north);
\node[lbl, above=0.7cm of cls] {$[:,0,:]$};
\node[lbl, above=0.7cm of tok] {$[:,1{:},:]$};
\node[lbl, above=0.7cm of msk] {\texttt{attn}$[:,1{:}]$};

\node[affinity, below=0.7cm of tok] (aff)
    {Shared affinity\quad
     \scriptsize$a_i = \dfrac{1}{\sqrt{H}}\,
       \mathbf{v}^{\!\top}\tanh\!\bigl(\mathbf{W}_c\mathbf{c}
       +\mathbf{W}_t\mathbf{h}_i\bigr)$};

\draw[arr] (cls.south) |- ([yshift=3pt]aff.west);
\draw[arr] (tok.south) -- (aff.north);

\node[dir1, below left=1.2cm and 0.3cm of aff] (d1)
    {\textbf{Dir 1:} \texttt{[CLS]} attends over tokens\\[1pt]
     \scriptsize$\boldsymbol{\alpha}=\mathrm{softmax}(\mathbf{a}\odot\mathbf{m}),\;
       \tilde{\mathbf{c}}=\boldsymbol{\alpha}\,\mathbf{T}$};
\node[dir2, below right=1.2cm and 0.3cm of aff] (d2)
    {\textbf{Dir 2:} tokens attend to \texttt{[CLS]}\\[1pt]
     \scriptsize$\boldsymbol{\beta}=\sigma(\mathbf{a})\odot\mathbf{m},\;
       \tilde{\mathbf{T}}=(\boldsymbol{\beta}/{\textstyle\sum}\beta_j)\,\mathbf{T}$};

\draw[arr] (aff.south) -- ++(0,-0.4) -| (d1.north);
\draw[arr] (aff.south) -- ++(0,-0.4) -| (d2.north);

\draw[darr] (msk.south) -- ++(0,-0.5) -| (d2.north)
    node[lbl, pos=0.62, right=4pt, fill=white]
    {\normalsize$\mathbf{m}$};
\draw[darr] (msk.south) -- ++(0,-2.4) -- ++(-6,0) |- (d1.east)
    node[lbl, pos=0.72, above=3pt, fill=white]
    {\normalsize$\mathbf{m}$};

\node[combine, below=2.6cm of aff] (lncat)
    {LayerNorm $+$ Concat\quad
     \scriptsize$\mathbf{f}=\bigl[\mathrm{LN}(\tilde{\mathbf{c}});\,
       \mathrm{LN}(\tilde{\mathbf{T}})\bigr]\in\mathbb{R}^{2H}$};

\draw[arr] (d1.south) |- ([yshift=-3pt]lncat.west);
\draw[arr] (d2.south) |- ([yshift=-3pt]lncat.east);

\node[mlp, below=0.6cm of lncat] (mlp)
    {MLP Classifier\quad
     \scriptsize Dropout $\to$ Linear$(2H\!\to\!H)$ $\to$
       GELU $\to$ Dropout $\to$ Linear$(H\!\to\!C)$};

\draw[arr] (lncat.south) -- (mlp.north);

\node[output, below=0.6cm of mlp] (out)
    {$\hat{y}\in\mathbb{R}^{C}$\quad(logits)};

\draw[arr] (mlp.south) -- (out.north);

\begin{scope}[on background layer]
  \node[draw=teal!50, dashed, rounded corners=5pt, fill=teal!4,
        fit=(cls)(tok)(msk), inner sep=5pt,
        label={[font=\scriptsize\color{teal!70!black}]above:encoder output split}
       ] {};
  \node[draw=violet!40, dashed, rounded corners=5pt, fill=violet!3,
        fit=(aff), inner sep=5pt] {};
  \node[draw=blue!35, dashed, rounded corners=5pt, fill=blue!3,
        fit=(d1), inner sep=5pt] {};
  \node[draw=orange!45, dashed, rounded corners=5pt, fill=orange!3,
        fit=(d2), inner sep=5pt] {};
\end{scope}
\end{tikzpicture}%
}
\caption{Dual pooling classification head on top of HelaBERT.
The encoder output is split into the \texttt{[CLS]} vector $\mathbf{c}$,
the token sequence $\mathbf{T}$, and a padding mask $\mathbf{m}$.
A shared affinity vector is computed via additive attention.
\textbf{Dir~1} (blue): \texttt{[CLS]} attends over tokens via softmax,
yielding $\tilde{\mathbf{c}}$.
\textbf{Dir~2} (orange): tokens attend back to \texttt{[CLS]} via a
sigmoid gate, yielding $\tilde{\mathbf{T}}$.
Both outputs are layer-normalized, concatenated to $\mathbb{R}^{2H}$,
and passed through a two-layer MLP to produce class logits.}
\label{fig:coattn-arch}
\end{figure*}
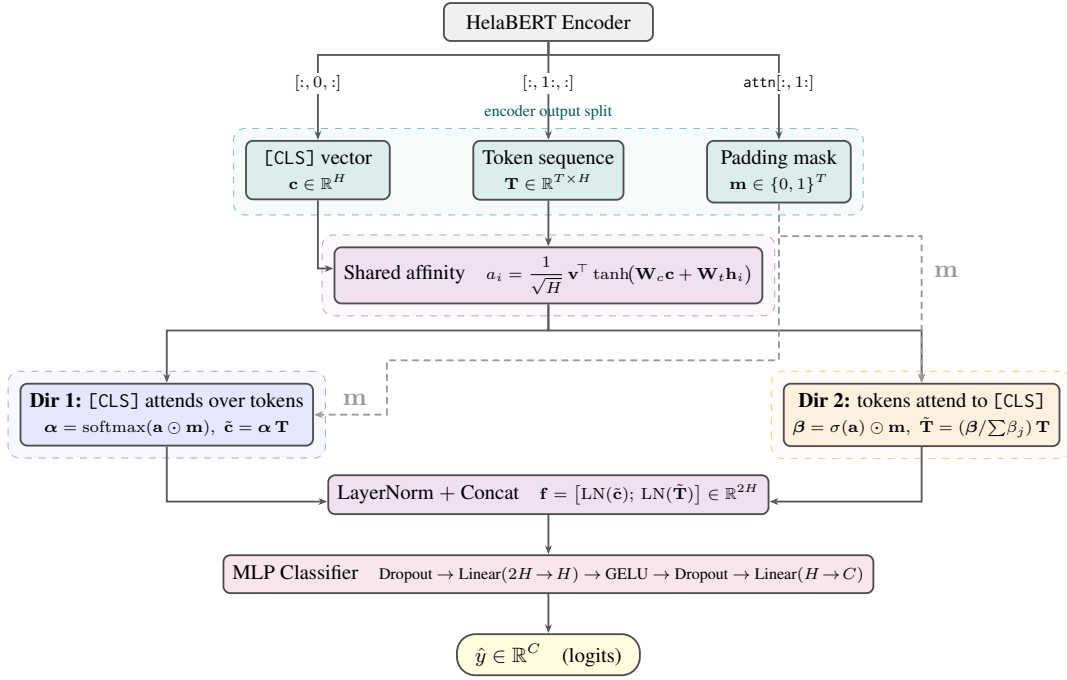

Figure~\ref{fig:coattn-arch} illustrates the full architecture.

\subsection{Training Configuration}
\label{subsec:coattn-training}

The dual pooling classification head uses the same 5-run seed protocol and 80/20
stratified split as the standard fine-tuning experiments
(Section~\ref{sec:training}). Hyperparameters are identical to those in
Table~\ref{tab:hyperparams} for each respective task.

\subsection{Results Across All Tasks}
\label{subsec:coattn-results}

Table~\ref{tab:coattn_all} summarises macro-F$_1$ for both the standard
and dual pooling classfication heads across all four tasks and both model sizes.
Detailed sentiment results including per-class scores are provided in
Tables~\ref{tab:coattn_sentiment} and~\ref{tab:coattn_perclass}.

\begin{table}[!htbp]
\centering
\small
\resizebox{\columnwidth}{!}{%
\begin{tabular}{llcccc}
\toprule
\textbf{Model} & \textbf{Head} &
\textbf{News Cat.} & \textbf{News Src.} &
\textbf{Sentiment} & \textbf{Writing} \\
\midrule
\multirow{3}{*}{Small}
  & Standard  & 85.97 & 60.16 & 65.34 & 95.92 \\
  & Co-attn   & 89.02 & 59.90 & 69.27 & 96.42 \\
  & $\Delta$  & \textbf{+3.05} & \textit{$-$0.26} & \textbf{+3.93} & +0.50 \\
\midrule
\multirow{3}{*}{Large}
  & Standard  & 90.38 & 63.65 & 64.60 & 97.73 \\
  & Co-attn   & 90.50 & 63.31 & 70.22 & 97.75 \\
  & $\Delta$  & +0.12 & \textit{$-$0.34} & \textbf{+5.62} & +0.02 \\
\bottomrule
\end{tabular}%
}
\caption{Standard \texttt{[CLS]}-linear head vs.\ dual pooling classification head:
macro-F$_1$ (\%) on all four tasks (best run of 5 seeds).
$\Delta$ = dual pooling $-$ standard. Positive $\Delta$ favours
dual pooling; negative values in \textit{italics}.}
\label{tab:coattn_all}
\end{table}

\begin{table}[!htbp]
\centering
\small
\resizebox{\columnwidth}{!}{%
\begin{tabular}{lcccc}
\toprule
\textbf{Model / Head} & \textbf{Acc.} & \textbf{Macro-F$_1$} & \textbf{W-F$_1$} \\
\midrule
HelaBERT-Small, standard   & 0.6589 & 0.6534 & 0.6581 \\
HelaBERT-Small, dual pooling    & 0.6881 & 0.6927 & 0.6904 \\
\quad$\Delta$ (dual pooling $-$ std.) & \textbf{+0.029} & \textbf{+0.039} & \textbf{+0.032} \\
\midrule
HelaBERT-Large, standard   & 0.6530 & 0.6460 & 0.6539 \\
HelaBERT-Large, dual pooling    & 0.7135 & 0.7022 & 0.7096 \\
\quad$\Delta$ (dual pooling $-$ std.) & \textbf{+0.061} & \textbf{+0.056} & \textbf{+0.056} \\
\bottomrule
\end{tabular}%
}
\caption{Standard vs.\ dual pooling sentiment results (best run of 5).
All figures on the same 3-class test set ($N=513$).}
\label{tab:coattn_sentiment}
\end{table}

\begin{table}[!htbp]
\centering
\small
\resizebox{\columnwidth}{!}{%
\begin{tabular}{lcccccc}
\toprule
& \multicolumn{3}{c}{\textbf{Standard}} & \multicolumn{3}{c}{\textbf{Dual pooling}} \\
\cmidrule(lr){2-4}\cmidrule(lr){5-7}
\textbf{Model} & NEG & NEU & POS & NEG & NEU & POS \\
\midrule
Small & 0.54 & 0.67 & 0.75 & 0.64 & 0.67 & 0.77 \\
Large & 0.55 & 0.68 & 0.70 & 0.59 & 0.73 & 0.78 \\
\bottomrule
\end{tabular}%
}
\caption{Per-class F$_1$ for standard vs.\ dual pooling on sentiment
(best run; $N=513$). NEG = negative, NEU = neutral, POS = positive.}
\label{tab:coattn_perclass}
\end{table}

\paragraph{Sentiment benefits most.}
The co-attention head yields the largest gains on sentiment:
$+3.9$ macro-F$_1$ points for HelaBERT-Small and $+5.6$ points for
HelaBERT-Large over the respective standard baselines. For
HelaBERT-Small, the gain is driven primarily by the \textsc{negative}
class, whose recall improves from 0.51 to 0.74 (+0.23) while F$_1$
increases from 0.54 to 0.64. The \textsc{neutral} and \textsc{positive}
classes are largely unaffected (+0.00 and +0.02 F$_1$ respectively).
For HelaBERT-Large, the pattern differs: \textsc{negative} recall
slightly decreases (0.55$\to$0.51) but precision gains 0.14 points
(0.55$\to$0.69), yielding a net F$_1$ improvement of $+0.04$.
More notably, the \textsc{neutral} class improves by $+0.05$ F$_1$
(0.68$\to$0.73), and \textsc{positive} improves by $+0.08$ (0.70$\to$0.78),
suggesting that the larger model's richer representations allow
co-attention to resolve ambiguous sentiment boundaries between
neutral and the other classes.

The larger absolute gain for HelaBERT-Large (+5.6 pp) relative to
HelaBERT-Small (+3.9 pp) is consistent with the intuition that richer
768-dimensional token representations provide more informative keys and
values for the co-attention computation, making the cross-interaction
between \texttt{[CLS]} and the token sequence more productive.

\paragraph{News category shows moderate gains for the small model.}
HelaBERT-Small improves by $+3.1$ points on news category, while
HelaBERT-Large gains only $+0.1$ points. We attribute this asymmetry
to representational capacity: at $H{=}384$, the \texttt{[CLS]} vector
provides a weaker sequence summary, and co-attention supplies a
meaningful second-order aggregation over the token sequence. At
$H{=}768$, the encoder's contextualised \texttt{[CLS]} representation
already captures sufficient category information, leaving little room
for the co-attention head to contribute further.

\paragraph{News source classification favours the standard head.}
The co-attention head performs marginally worse on news source
classification for both model sizes ($\sim$0.3 pp for Small,
$\sim$0.3 pp for Large). News source headlines average only $\sim$8
tokens --- the shortest inputs across all four tasks. With so few
real tokens, the attended representations in both co-attention directions
are computed over a very sparse key set, providing limited additional
signal beyond the \texttt{[CLS]} token itself. Furthermore, the co-attention
head introduces approximately $4H^2$ additional parameters
(${\approx}590$K for Small, ${\approx}2.4$M for Large), which 3 training
epochs over short headlines is insufficient to fully converge.

\paragraph{Writing style gains are negligible.}
Both models show near-zero macro-F$_1$ gains on writing style
($+0.5$ pp for Small, $+0.0$ pp for Large). This task operates at
performance ceiling ($>$95\% macro-F$_1$), leaving little measurable
headroom for any architectural improvement.

\begin{figure}[t]
  \centering
  \includegraphics[width=\columnwidth]{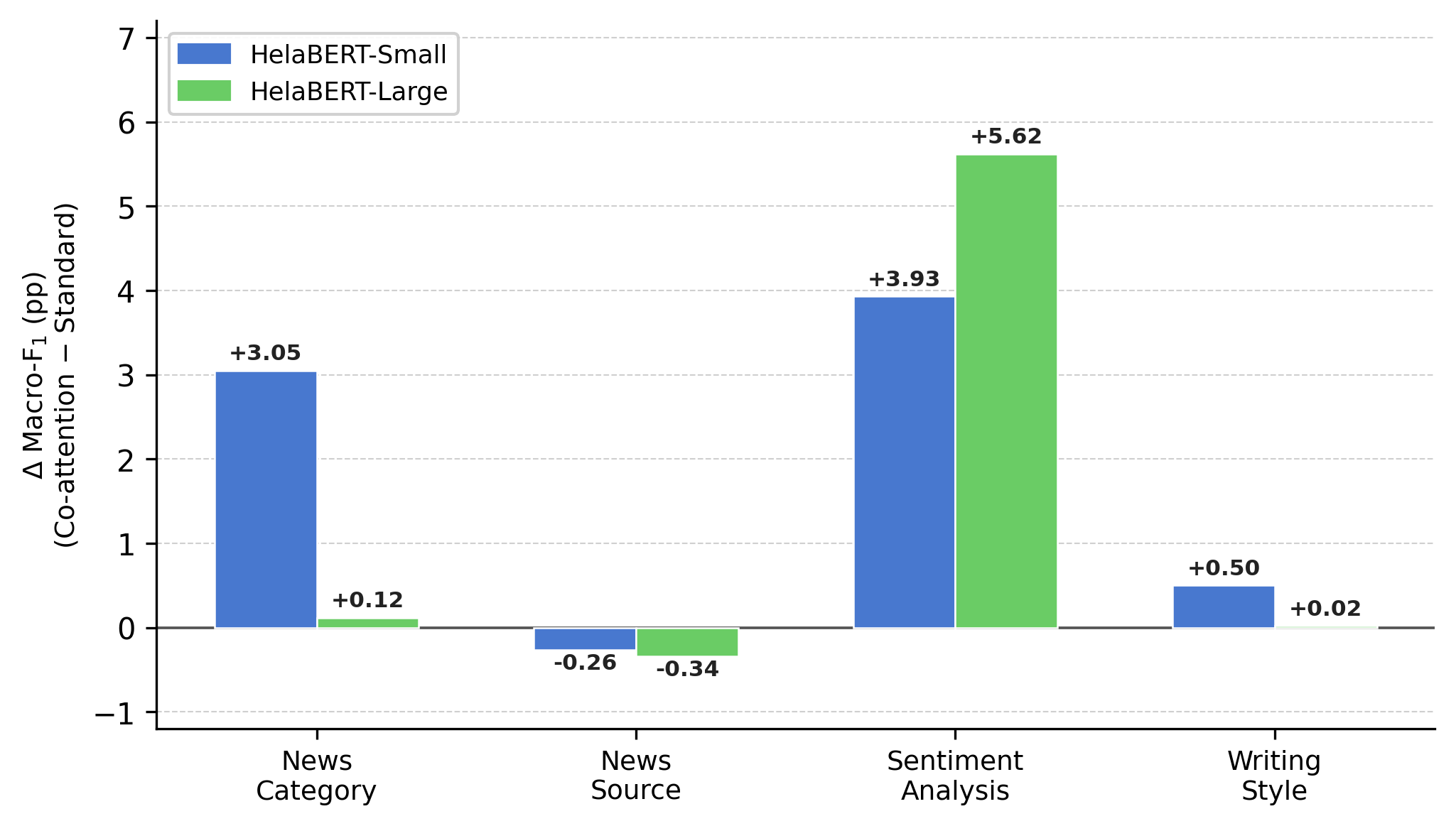}
  \caption{Co-attention vs.\ standard head: $\Delta$ macro-F$_1$
           (percentage points) across all four tasks for HelaBERT-Small
           (blue) and HelaBERT-Large (green). Bars below zero indicate
           tasks where the standard head outperforms co-attention.}
  \label{fig:coattn-delta}
\end{figure}

\paragraph{Summary.}
Taken together, these results suggest that co-attention classification
heads are most beneficial when the task relies on a small number of
discriminative tokens within moderate-length sequences, and when the
base model's representational capacity is limited. For very short
sequences ($\lesssim$8 tokens), near-saturated tasks, or large models
with strong \texttt{[CLS]} representations, the standard \texttt{[CLS]}-linear
head remains competitive. Figure~\ref{fig:coattn-delta} summarises the
$\Delta$ macro-F$_1$ gains across all tasks and both model sizes.

\section{Discussion}
\label{sec:discussion}

\paragraph{HelaBERT vs.\ multilingual and monolingual baselines.}
HelaBERT-Large surpasses XLM-R-large on news category classification
despite Sinhala comprising only $\sim$0.15\% of XLM-R's pre-training
data, confirming that a monolingual model on a focused corpus can
overcome multilingual capacity dilution. It also outperforms
SinBERT-Large on all four tasks despite SinBERT using the stronger
RoBERTa recipe. We attribute this to HelaBERT's substantially larger
corpus ($\sim$1.1B vs.\ $\sim$192M tokens) and its Sinhala-specific
SentencePiece Unigram tokenizer, which better handles agglutinative
morphology than BPE trained on a multilingual vocabulary.
The only task where HelaBERT-Large trails XLM-R-large is writing
style ($-$0.7 pp), likely due to XLM-R's cross-lingual exposure to
diverse document styles providing complementary inductive biases.

\paragraph{When does co-attention help?}
Co-attention is most effective when polarity or category is signaled
by a few discriminative tokens within moderate-length sequences and
when encoder capacity is limited. Sentiment benefits most (+3.9--5.6
pp) because negations and intensifiers are sparse yet decisive.
Gains are negligible on writing style (near ceiling, $>$95\%) and
slightly negative on news source ($\sim$8-token headlines too short
for meaningful attention over tokens). The larger gain for
HelaBERT-Large on sentiment (+5.6 pp vs.\ +3.9 pp) suggests richer
768-dimensional representations make co-attention more discriminative,
indicating a productive interaction between model scale and the
proposed head.

\section{Limitations}
\label{sec:limitations}

Both HelaBERT models are monolingual and do not support cross-lingual
transfer. The SentencePiece tokenizer requires manual loading via the
\texttt{sentencepiece} library and is not compatible with the HuggingFace
\texttt{AutoTokenizer} API out of the box. HelaBERT-Small was pre-trained
on 256-token windows despite a 512-token positional limit; performance on
sequences longer than 256 tokens is therefore untested. The web-crawled
pre-training corpus may retain noise not fully removed by preprocessing
and skews toward formal written Sinhala, potentially underrepresenting
dialectal and colloquial registers. Finally, downstream evaluation is
restricted to text classification; sequence labelling, question answering,
and generative tasks are left to future work.

\section{Conclusion}
\label{sec:conclusion}

We presented HelaBERT, a family of two BERT-based masked language
models pre-trained from scratch on Sinhala text. HelaBERT-Large
achieves state-of-the-art results among monolingual Sinhala models on
all four evaluated classification tasks, surpassing XLM-R-large on
news category classification, news source classification and outperforming SinBERT-Large across the board. HelaBERT-Small provides a competitive lightweight
alternative with substantially fewer parameters.

We proposed and systematically evaluated a co-attention classification
head that computes bidirectional attention between the \texttt{[CLS]}
token and the full token sequence. The head yields consistent gains on
sentiment analysis (+3.9--5.6 macro-F$_1$ points) and a moderate
improvement on news category classification for HelaBERT-Small
(+3.1 points), while the standard \texttt{[CLS]}-linear head remains
competitive on short-input and near-saturated tasks.

Both models and the SentencePiece Unigram tokenizer are released
publicly to support further research in Sinhala NLP. Future work
includes extending evaluation to sequence labelling and question
answering tasks, exploring continued pre-training with additional
Sinhala data, and investigating the co-attention mechanism in
combination with larger model architectures.

\bibliography{custom}
 
\appendix
 
\section{Pre-training Hyperparameters}
\label{sec:appendix-hyperparams}
 
Table~\ref{tab:pretrain-config} lists the full set of pre-training
hyperparameters for HelaBERT-Small and HelaBERT-Large, referenced in
Section~\ref{subsec:pretraining-config}.
 
\begin{table}[!htbp]
\centering
\small
\resizebox{\columnwidth}{!}{%
\begin{tabular}{lll}
\toprule
\textbf{Hyperparameter} & \textbf{Small} & \textbf{Large} \\
\midrule
Sequence length             & 256 (stride 128) & 512             \\
Total training samples      & $\sim$7.4M       & $\sim$4.3M      \\
Train / validation split    & 90\% / 10\%      & 90\% / 10\%     \\
MLM probability             & 15\%             & 15\%            \\
Per-device batch size       & 32               & 256             \\
Gradient accumulation steps & 8                & 1               \\
Effective batch size        & 256              & 256             \\
Learning rate               & $1\times10^{-4}$ & $1\times10^{-4}$\\
LR scheduler                & Cosine           & Cosine          \\
Warmup ratio                & 5\%              & 10\%            \\
Weight decay                & 0.01             & 0.01            \\
Epochs                      & 2                & 6               \\
Mixed precision             & FP16             & BF16            \\
\bottomrule
\end{tabular}%
}
\caption{Pre-training hyperparameters for HelaBERT-Small and HelaBERT-Large.}
\label{tab:pretrain-config}
\end{table}

\section{Dataset Details}
\label{sec:appendix-datasets}
 
This appendix provides exhaustive details for the four fine-tuning
datasets summarised in Table~\ref{tab:datasets}
(Section~\ref{sec:finetuning-datasets}), including per-class label
distributions, licensing and access information, and collection
methodology.
 
\subsection{News Category Classification}
\label{subsec:appendix-newscat}
 
Table~\ref{tab:appendix-newscat} reports the per-class train/test split
for the five news category labels. Label identities in the released
dataset are numeric (0--4); the original dataset card does not document
a mapping from these numeric IDs to the five named categories used in
the main text (\textit{political}, \textit{business},
\textit{technology}, \textit{sports}, \textit{entertainment}), so we
report counts by numeric label rather than assume a correspondence.
 
\begin{table}[!htbp]
\centering
\small
\begin{tabular}{lrr}
\toprule
\textbf{Label} & \textbf{Train} & \textbf{Test} \\
\midrule
0     & 418  & 102 \\
1     & 365  & 91  \\
2     & 675  & 166 \\
3     & 797  & 199 \\
4     & 341  & 82  \\
\midrule
Total & 2{,}596 & 640 \\
\bottomrule
\end{tabular}
\caption{Per-label train/test distribution for News Category
Classification.}
\label{tab:appendix-newscat}
\end{table}
 
 
\subsection{News Source Classification}
\label{subsec:appendix-newssource}
 
Table~\ref{tab:appendix-newssource} reports the per-source train/test
split across the nine news source labels. As with the category
dataset, source identities are released only as numeric labels
(0--8); the dataset card does not document a mapping to the underlying
site names, so we report counts by numeric label.
 
\begin{table}[!htbp]
\centering
\small
\begin{tabular}{lrr}
\toprule
\textbf{Label} & \textbf{Train} & \textbf{Test} \\
\midrule
0     & 2{,}233  & 558  \\
1     & 1{,}303  & 326  \\
2     & 2{,}334  & 584  \\
3     & 1{,}143  & 286  \\
4     & 2{,}366  & 592  \\
5     & 2{,}282  & 571  \\
6     & 2{,}243  & 560  \\
7     & 2{,}201  & 550  \\
8     & 2{,}175  & 544  \\
\midrule
Total & 18{,}280 & 4{,}571 \\
\bottomrule
\end{tabular}
\caption{Per-label train/test distribution for News Source
Classification.}
\label{tab:appendix-newssource}
\end{table}
 
 
\subsection{Sentiment Analysis}
\label{subsec:appendix-sentiment}
 
Table~\ref{tab:appendix-sentiment} reports the per-class train/test
split for the three-class sentiment dataset used in this work (see
Section~\ref{subsec:sentiment-dataset} for discussion of why this
differs from the four-class dataset used by prior baselines).
 
\begin{table}[!htbp]
\centering
\small
\begin{tabular}{lrr}
\toprule
\textbf{Label} & \textbf{Train} & \textbf{Test} \\
\midrule
Neutral  & 899  & 225 \\
Positive & 599  & 150 \\
Negative & 551  & 138 \\
\midrule
Total    & 2{,}049 & 513 \\
\bottomrule
\end{tabular}
\caption{Per-label train/test distribution for Sentiment Analysis.}
\label{tab:appendix-sentiment}
\end{table}
 
 
\subsection{Writing Style Classification}
\label{subsec:appendix-writingstyle}
 
Table~\ref{tab:appendix-writingstyle} reports the per-style train/test
split for the four writing style labels.
 
\begin{table}[!htbp]
\centering
\small
\begin{tabular}{lrr}
\toprule
\textbf{Label} & \textbf{Train} & \textbf{Test} \\
\midrule
News     & 3{,}568 & 885 \\
Academic & 2{,}834 & 701 \\
Creative & 1{,}919 & 458 \\
Blog     & 1{,}687 & 418 \\
\midrule
Total    & 10{,}008 & 2{,}462 \\
\bottomrule
\end{tabular}
\caption{Per-label train/test distribution for Writing Style
Classification.}
\label{tab:appendix-writingstyle}
\end{table}
 
 
\end{document}